\documentclass[runningheads]{llncs}
\usepackage{graphicx}
\usepackage{todonotes}
\usepackage{multirow}

\usepackage{array}
\usepackage{tabu}
\newcolumntype{M}[1]{>{\centering\arraybackslash}m{#1}}
\newcolumntype{L}[1]{>{\arraybackslash}m{#1}}

\begin{document}
\title{Overview of BioASQ 2021: The ninth BioASQ challenge on Large-Scale Biomedical Semantic Indexing and Question Answering
}
    \titlerunning{Overview of BioASQ 2021}
%

\author{
Anastasios Nentidis\inst{1,2} \and
Georgios Katsimpras\inst{1} \and
Eirini Vandorou\inst{1} \and
Anastasia Krithara\inst{1} \and 
Luis Gasco\inst{3} \and
Martin Krallinger\inst{3} \and
Georgios Paliouras\inst{1}
}
\authorrunning{A. Nentidis et al.}
%
\institute{
National Center for Scientific Research ``Demokritos'', Athens, Greece\\
\email{\{tasosnent, gkatsibras, evandorou, akrithara, paliourg\}@iit.demokritos.gr}\\
\and
Aristotle University of Thessaloniki, Thessaloniki, Greece\\ \and
Barcelona Supercomputing Center, Barcelona, Spain\\
\email{\{martin.krallinger, lgasco\}@bsc.es}
}
\maketitle              
\begin{abstract}
Advancing the state-of-the-art in large-scale biomedical semantic indexing and question answering is the main focus of the BioASQ challenge. BioASQ organizes respective tasks where different teams develop systems that are evaluated on the same benchmark datasets that represent the real information needs of experts in the biomedical domain. This paper presents an overview of the ninth edition of the BioASQ challenge in the context of the Conference and Labs of the Evaluation Forum (CLEF) 2021. 
In this year, a new question answering task, named Synergy, is introduced to support researchers studying the COVID-19 disease and measure the ability of the participating teams to  discern information while the problem is still developing. In total, 42 teams with more than 170 systems were registered to participate in the four tasks of the challenge. The evaluation results, similarly to previous years, show a performance gain against the baselines which indicates the continuous improvement of the state-of-the-art in this field.


\end{abstract}
\section{Introduction}
In this paper, we present the shared tasks and the datasets of the ninth BioASQ challenge in 2021, as well as we as an overview of the participating systems and their performance.
The remainder of this paper is organized as follows. Section~\ref{sec:tasks} provides an overview of the shared tasks, that took place from December 2020 to May 2021, and the corresponding datasets developed for the challenge. 
Section~\ref{sec:participants} presents a brief overview of the systems developed by the participating teams for the different tasks. 
Detailed descriptions for some of the systems are available in the proceedings of the lab. 
Then, in section~\ref{sec:results}, we focus on evaluating the performance of the systems for each task and sub-task, using state-of-the-art evaluation measures or manual assessment.
Finally, section~\ref{sec:conclusion} draws some conclusions regarding this version  of the BioASQ challenge.

\section{Overview of the Tasks}
\label{sec:tasks}
In this year, the ninth version of the BioASQ challenge offered four tasks: (1) a large-scale
biomedical semantic indexing task (task 9a), (2) a biomedical question answering task (task 9b), both considering documents in English, (3) a medical semantic indexing in Spanish (task MESINESP9 using literature, patents and clinical trial abstracts), and (4) a new task on biomedical question answering on the developing problem of COVID-19 (task Synergy). In this section, we describe the two established tasks 9a and 9b with focus on differences from previous versions of the challenge~\cite{nentidis2020overview}. Detailed information about these tasks can be found in \cite{Tsatsaronis2015}. Additionally, we discuss the second version of the MESINESP task and also present the new Synergy task on biomedical question answering for developing problems, which was introduced this year, providing statistics about the dataset developed for each task.

\subsection{Large-scale semantic indexing - Task 9a}

\begin{table}[!htb]
        \centering
    \begin{tabular}{M{0.1\linewidth}M{0.15\linewidth}M{0.3\linewidth}M{0.3\linewidth}}\hline
        \textbf{Batch} & \textbf{Articles} & \textbf{Annotated Articles} & \textbf{Labels per Article}  \\ \hline
        \multirow{5}{*}{1}        & 7967       & 7808                          & 12.61                           \\
                              & 10053       & 9987                          & 12.40                             \\
                              & 4870      & 4854                         & 12.16                            \\
                              & 5758       & 5735                          & 12.34                             \\
                              & 5770       & 5666                           & 12.49                             \\ \hline

    Total                  & 34418      & 34050                         & 12.42                             \\ \hline
    \multirow{5}{*}{2}        & 6376       & 6374                          & 12.39                             \\
                              & 9101       & 6403                          & 11.76                             \\
                              & 7013       & 6590                          & 12.15                             \\
                              & 6070       & 5914                          & 12.62                             \\
                              & 6151       & 5904                          & 12.63                             \\  \hline

    Total                  & 34711      & 31185                         & 12.30                              \\ \hline
    \multirow{5}{*}{3}        & 5890       & 5730                          & 12.81                             \\
                              & 10818       & 9910                          & 13.03                             \\
                              & 4022       & 3493                          & 12.21                            \\
                              & 5373       & 4005                          & 12.62                             \\
                              & 5325       & 2351                          & 12.97                             \\ \hline
    Total                  & 31428      & 25489                         & 12.71 \\  \hline
    
    \end{tabular}
        \caption{Statistics on test datasets for Task 9a.}\label{tab:a_data}
\end{table}

The aim of Task 9a is to classify articles from the PubMed/MedLine\footnote{https://pubmed.ncbi.nlm.nih.gov/} digital library into concepts of the MeSH hierarchy. Specifically, the test sets for the evaluation of the competing systems consist of new PubMed articles that are not yet annotated by the indexers in the National Library of Medicine (NLM). Table \ref{tab:a_data} illustrates a more detailed view of each test.
As in the previous years, the task is realized in three independent runs of 5 weekly test sets each. Two scenarios are provided: i) on-line and ii) large-scale. The test sets are a collection of new articles without any restriction on the journal published. 
For the evaluation of the competing systems standard flat information retrieval measures are used, as well as hierarchical ones, comparing the predictions of the participants with the annotations from the NLM indexers, once available.
Similarly to the previous years, for each test set, participants are required to submit their answers in 21 hours. 
Furthermore, a training dataset was available for Task 9a which contains 15,559,157 articles with 12.68 labels per article, on average, and covers 29,369 distinct MeSH labels in total.

\subsection{Biomedical semantic QA - Task 9b}

Task 9b focuses on enabling the competing teams to develop systems for all the stages of question answering in the biomedical domain by introducing a large-scale question answering challenge. Again this year, four types of questions are considered: “yes/no”, “factoid”, “list” and “summary” questions \cite{balikas13}.
A total of 3,743 questions, which are annotated with golden relevant elements and answers from previous versions of the task, consist of the available training dataset for this task. The dataset is used by the participating teams to develop their systems. Table \ref{tab:b_data} provides detailed information about both training and testing sets.

\begin{table}[!htb]
        \centering
        \begin{tabular}{M{0.1\linewidth}M{0.08\linewidth}M{0.08\linewidth}M{0.12\linewidth}M{0.1\linewidth}M{0.15\linewidth}M{0.15\linewidth}M{0.15\linewidth}}\hline
        \textbf{Batch} 	& \textbf{Size} 	&	\textbf{Yes/No}	&\textbf{List}	&\textbf{Factoid}	&\textbf{Summary}& \textbf{Documents} 	& \textbf{Snippets}  	\\ \hline
        Train       &		3,743		&	1033	&719	&1092		&899	&		9.43			&	12.32			 	\\
        Test 1		&		100			&	27		&21		&29			&23		&		3.40			&	4.66			 	\\
        Test 2		&		100			&	22		&20		&34			&24		&		3.43			&	4.88			 	\\
        Test 3		&		100			&	26		&19		&37			&18		&		3.21			&	4.29			 	\\ 
        Test 4		&		100			&	25		&19		&28			&28		&		3.10			&	4.01			 	\\
        Test 5		&		100			&	19		&18		&36			&27		&		3.59			&	4.69			 	\\ \hline                    
        \textbf{Total}	&	4,243		&	1152	&816	&1256		&1019	&		8.71			&	11.40			 	\\ \hline 
        
        \end{tabular}
        \caption{Statistics on the training and test datasets of Task 9b. The numbers for the documents and snippets refer to averages per question.}\label{tab:b_data}
\end{table}

Task 9b is divided into two phases: (phase A) the retrieval of the required information and (phase B) answering the question. Moreover, it is split into five independent bi-weekly batches and the two phases for each batch run during two consecutive days. In each phase, the participants receive the corresponding test set and have 24 hours to submit the answers of their systems.
More precisely, in phase A, a test set of 100 questions written in English is released and the participants are expected to identify and submit relevant elements from designated resources, including PubMed/MedLine articles, snippets extracted from these articles, concepts and RDF triples. 
In phase B, the manually selected relevant articles and snippets for these 100 questions are also released and the participating systems are asked to respond with \textit{exact answers}, that is entity names or short phrases, and \textit{ideal answers}, that is natural language summaries of the requested information.

\subsection{Medical semantic indexing in Spanish - MESINESP}

Over the last year, scientific production has increased significantly and has made more evident than ever the need to improve the information retrieval methods under a multilingual IR or search scenario for medical content beyond data only in English \cite{torres2021growth}. The scenario faced during the year 2020 demonstrates the need to improve access to information in demanding scenarios such as a disease outbreaks or public health threats at multinational/cross-border scale. In a health emergency scenario, access to scientific information is essential to accelerate research and healthcare progress and to enable resolving the health crisis more effectively. During the COVID-19 health crisis, the need to improve multilingual search systems became evermore significant, since a considerable fraction of medical publications (especially clinical case reports on COVID patients) were written in the native language of medical professionals. 

MESINESP was created in response to the lack of resources for indexing content in languages other than English, and to improve the lack of semantic interoperability in the search process when attempting to retrieve medically relevant information across different data sources.

The MESINESP 2021 track \cite{luis2020overview}, promoted by the Spanish Plan for the Advancement of Language Technology (Plan TL) \footnote{https://plantl.mineco.gob.es} and organized by the Barcelona Supercomputing Center (BSC) in collaboration with BioASQ, aims to improve the state of the art of semantic indexing for content written in Spanish, ranking among the highest number of native speakers in the world \footnote{https://www.ethnologue.com/guides/ethnologue200}. In an effort to improve interoperability in semantic search queries, this edition was divided into three subtracks to index scientific literature, clinical trials and medical patents.

\begin{figure*}[!htb]
\centerline{\includegraphics[width=1\textwidth]{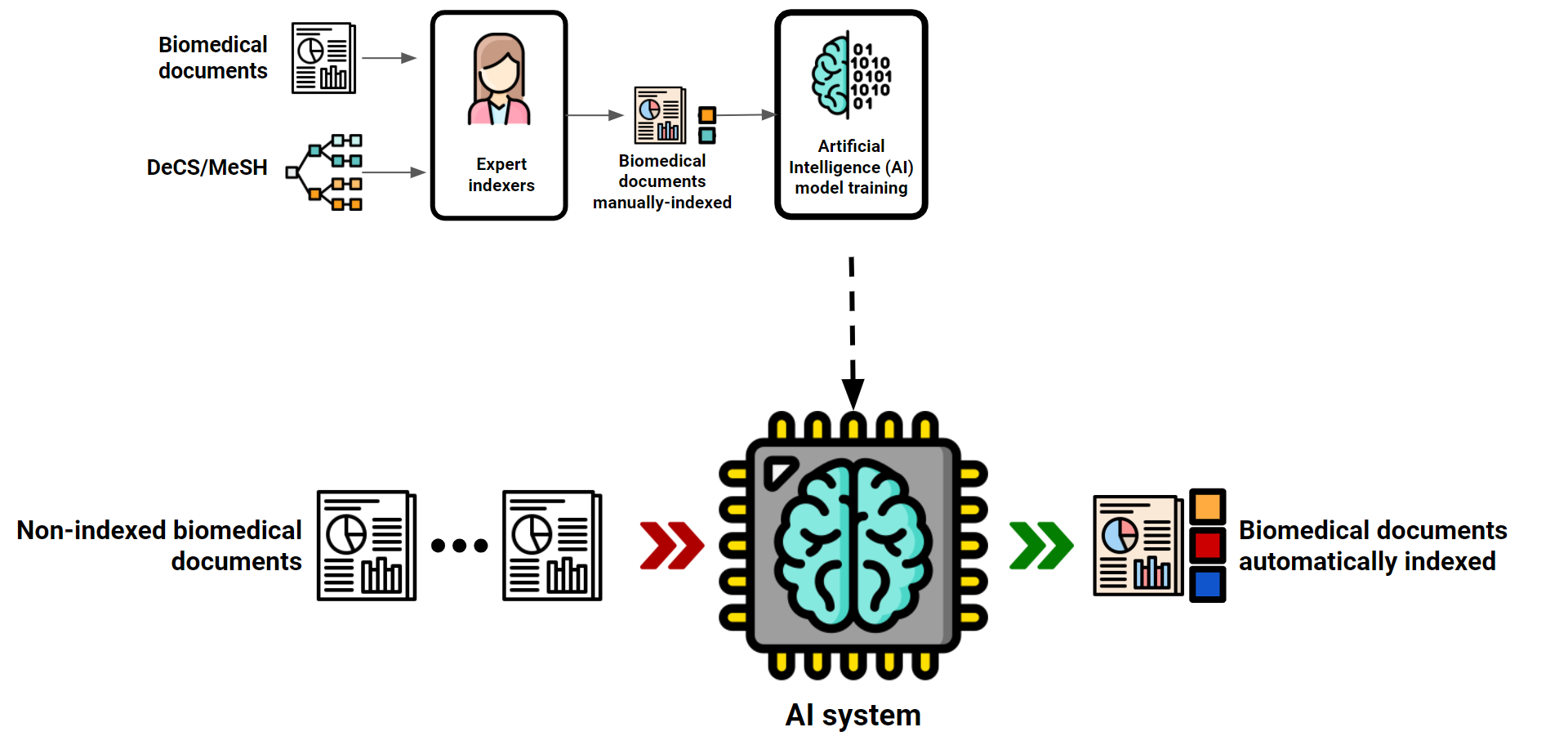}}
\caption{Simplified MESINESP2 workflow showing the importance of annotation for the generation of automatic semantic indexing systems.}\label{fig:02}
\end{figure*}

MESINESP-L (subtrack 1)  required the automatic indexing with DeCS\footnote{DeCS (\textit{Descriptores Descriptores en Ciencias de la Salud}, Health Science Descriptors) is a structured controlled vocabulary created by BIREME to index scientific publications on BvSalud (\textit{Biblioteca Virtual en Salud}, Virtual Health Library)} terms of a set of abstracts from scientific articles (titles and abstracts) from two widely used literature databases with content in Spanish: IBECS\footnote{IBECS includes bibliographic references from scientific articles in health sciences published in Spanish medical journals. http://ibecs.isciii.es} and LILACS\footnote{LILACS is a resource comprising scientific and technical literature from Latin America and the Caribbean countries. It includes 26 countries, 882 journals and 878,285 records, 464,451 of which are full texts https://lilacs.bvsalud.org}. 

We built the corpora for the task from the data available in BvSalud, the largest database of scientific documents in Spanish, which integrates records from LILACS, MEDLINE, IBECS and other databases. First, we downloaded the whole collection of 1.14 million articles present in the platform. Then, only journal articles with titles and abstracts written in Spanish that had been previously manually indexed by LILACS and IBECS experts with DeCS codes were selected, obtaining a final training dataset of 237,574 articles. A development set of records manually indexed by expert annotators was also provided. This development corpus included 1,065 articles manually annotated (indexed) by the three human indexers who obtained the best IAA in the last MESINESP edition. To generate the test set, 500 publications were selected to be indexed by the three experts. We also incorporated a background set of 9,676 Spanish-language clinical practice guidelines to evaluate the performance of the models on this type of biomedical documents.

Clinical Trials subtrack (MESINESP-T) asked participating teams to generate models able to automatically predict DeCS codes for clinical trials from the REEC database\footnote{ Registro Español de Estudios Clínicos, a database containing summaries of clinical trials https://reec.aemps.es/reec/public/web.html}.  

Last year's task generated a silver standard (automatically assigned codes by participating teams) with a set of REEC clinical trials. The predictions of the best performing team was used as a substitute or surrogate data collection for training systems, pooling a total of 3,560 clinical trials. For the development set, 147 records manually annotated by expert indexers in MESINESP 2020 were provided. For the test set, we calculated the semantic similarity between MESINESP-L training corpus and a pre-selection of 416 clinical trials published after 2020. Then, the top 250 most similar clinical trials, which included many COVID-19 related trials, were annotated by our indexers. Similar to what was done for the scientific literature track, we included a background set of 5,669 documents from medicine data sheets to be automatically indexed by teams (generating thus a silver standard collection).

Finally, for the patents subtrack (MESINESP-P), the aim was to explore and evaluate indexing strategies of medical patents written in Spanish providing only a very small manually annotated patent collection (in addition to the literature corpus). We presented the track as a cross-corpus training challenge, in which participants should transfer/adapt previous models to the patent langaguage without a large manually annotated data set.  All patents written in Spanish having the assigned IPC codes "A61P" and "A61K31" were retrieved using Google Big Query\footnote{https://cloud.google.com/blog/topics/public-datasets/google-patents-public-datasets-connecting-public-paid-and-private-patent-data}, only these codes were considered as they cover medicinal chemistry related topics \cite{krallinger2015overview}. After data harvesting, 65,513 patents were obtained, out of which the 228 most semantically similar to the MESINESP-L training set were chosen. After an annotation process, 119 were used as the development set and 109 as the test set. Some summary statistics of the used datasets can be seen in the Table \ref{tab:stats_corpus}

\begin{table}[]
\resizebox{\textwidth}{!}{%
\begin{tabular}{@{}lcccccc@{}}
\toprule
\multicolumn{1}{c}{\textbf{Corpus}} &
  \multicolumn{1}{l}{\textbf{Docs}} &
  \multicolumn{1}{l}{\textbf{DeCS}} &
  \multicolumn{1}{l}{\textbf{Unique DeCS}} &
  \multicolumn{1}{l}{\textbf{Tokens}} &
  \multicolumn{1}{l}{\textbf{Avg.DeCS/doc}} &
  \multicolumn{1}{l}{\textbf{Avg.token/doc}} \\ \midrule
\textbf{MESINESP-L Training}   & 237574 & $\sim$1.98M & 22434 & $\sim$43.1M & 8.37(3.5)   & 181.45(72.3)    \\
\textbf{MESINESP-L Development} & 1065   & 11283       & 3750  & 211420      & 10.59(4.1)  & 198.52(64.2)    \\
\textbf{MESINESP-L Test}         & 491    & 5398        & 2124  & 93645       & 10.99(3.9)  & 190.72(63.6)    \\ \midrule
\textbf{MESINESP-T Training}     & 3560   & 52257       & 3940  & $\sim$4.13M & 14.68(1.19) & 1161.0(553.5)   \\
\textbf{MESINESP-T Development} & 147    & 2038        & 771   & 146791      & 13.86(5.53) & 998.58(637.5)   \\
\textbf{MESINESP-T Test}         & 248    & 3271        & 905   & 267031      & 13.19(4.49) & 1076.74(553.68) \\ \midrule
\textbf{MESINESP-P Development}  & 109    & 1092        & 520   & 38564       & 10.02(3.11) & 353.79(321.5)   \\
\textbf{MESINESP-P Test}         & 119    & 1176        & 629   & 9065        & 9.88(2.76)  & 76.17(27.36)    \\ \bottomrule
\end{tabular}%
}
\caption{Summary statistics of the MESINESP corpora.}
\label{tab:stats_corpus}

\centering
\begin{tabular}{@{}ccccc@{}}
\toprule
\textbf{Corpus} &
  \textbf{Diseases} &
  \textbf{Medications} &
  \textbf{Procedures} &
  \textbf{Symptoms} \\ \midrule
\textbf{MESINESP - L} & 711751 & 87150 & 362927 & 127810 \\
\textbf{MESINESP - T} & 129362 & 86303 & 52566  & 10140  \\
\textbf{MESINESP - P} & 171    & 180   & 25     & 12     \\ \bottomrule
\end{tabular}
\caption{Summary statistics on the number of entities of each type extracted for each corpus.}
\end{table}

Some additional resources were published in order to serve as complementary annotations for participating teams. Since the BSC text mining unit had already implemented several competitive medical named entity recognition tools adapted to content in Spanish ~\cite{miranda2021profner,miranda2020overview,miranda2020named}, four different NER systems were applied to each of the corpora to annotate automatically mentions of medical entities that may help improve model performance, namely diseases, procedures, medications/drugs and symptoms. Many DeCS terms do actually correspond to these semantic classes, in particular diseases. Overall, the semantic annotation results for the MESINESP included around 840,000 disease mentions, 170,000 medicine/drug mentions, 415,000 medical procedures mentions and 137,000 symptoms mentions.

\subsection{Synergy Task}

The established question answering BioASQ task (Task B) is structured in a sequence of phases. First comes the annotation phase; then with a partial overlap runs the challenge; and only when this is finished does the assessment phase start. 
This leads to restricted interaction between the participating systems and the experts, which is acceptable due to the nature of the questions, that have a clear, undisputed answer.
However, a more interactive model is necessary for open questions on developing research topics, such as the case of COVID-19, where new issues appear every day and most of them remain open for some time.
In this context, a model based on the synergy between the biomedical experts and the automated question answering systems is needed.


 In this direction, we introduced the BioASQ Synergy task envisioning a continuous dialog between the experts and the systems. In this model, the experts pose open questions and the systems provide relevant material and answers for these questions. Then, the experts assess the submitted material (documents and snippets) and answers, and provide feedback to the systems, so that they can improve their responses. 
 This process proceeds with new feedback and new predictions from the systems in an iterative way. 
 
 This year, Task Synergy took place in two versions, focusing on unanswered questions for the developing problem of the COVID-19 disease.
 Each version was structured into four rounds, of systems responses and expert feedback for the same questions. However, some new questions or new modified versions of some questions could be added to the test sets. 
 The details of the datasets used in task Synergy are available in Table \ref{tab:syn_data}.

  \begin{table}[!htb]
        \centering
        \begin{tabular}{c c c c c c c c c}\hline
      	\textbf{Version}  & \textbf{Round}  & \textbf{Size}  & \textbf{Yes/No} &\textbf{List} &\textbf{Factoid} &\textbf{Summary}& \textbf{Answer}  & \textbf{Feedback}   \\
		\hline
		 
	1	&  1  &  108   & 33  &22  &17 &36  &	 0  & 0 \\
	1	&  2  &  113   & 34  &25  &18 &36  &  53   & 101  \\
	1	&  3  &  113   & 34  &25  &18 &36  &  80   & 97 \\ 
	1	&  4  &  113   & 34  &25  &18 &36  &  86  & 103  \\
	\hline
	2	&  1  &  95   & 31  &22  &18  &24  & 6  & 95 \\
	2	&  2  &  90   & 27  &22  &18  &23  &  10 & 90 \\
	2	&  3  &  66  & 17 & 14 & 18 & 17 & 25 & 66 \\
	2	&  4  &  63  & 15 & 14 & 17 & 17 & 33 & 63 \\
	 \hline                 
	\end{tabular}
	\caption{Statistics on the datasets of Task Synergy. ``Answer'' stands for questions marked as having enough relevant material from previous rounds to be answered".}\label{tab:syn_data}
\end{table}
 
 Contrary to the task B, this task was not structured into phases, but both relevant material and answers were received together. 
 However, for new questions only relevant material (documents and snippets) is required until the expert considers that enough material has been gathered during the previous round and mark the questions as ``ready to answer". 
 When a question receives a satisfactory answer that is not expected to change, the expert can mark the question as ``closed", indicating that no more material and answers are needed for it. 

In each round of this task, we consider material from the current version of the COVID-19 Open Research Dataset (CORD-19) \cite{wang2020cord} to reflect the rapid developments in the field.
As in task B, four types of questions are supported, namely yes/no, factoid, list, and summary, and two types of answers, exact and ideal.
The evaluation of the systems will be based on the measures used in Task 9b. Nevertheless, for the information retrieval part we focus on new material. Therefore, material already assessed in previous rounds, available in the expert feedback, should not be re-submitted.
Overall, through this process, we aim to facilitate the incremental understanding of COVID-19 and contribute to the discovery of new solutions.  


\section{Overview of participation}
\label{sec:participants}
\subsection{Task 9a}
This year, 6 teams participated with a total of 21 different systems. Below, we provide a brief overview of those systems for which a description was available,
stressing their key characteristics. The participating systems along with their corresponding approaches are listed in Table~\ref{tab:a_sys}. Detailed descriptions for some of the systems are available at the proceedings of the workshop.

\begin{table}[!htb]
        \centering
        \begin{tabular}{M{0.2\linewidth}M{0.8\linewidth}}\hline
        \textbf{System} & \textbf{Approach} \\ \hline
        bert\_dna, pi\_dna & SentencePiece, BioBERT, multiple binary classifiers\\\hline
        NLM & SentencePiece, CNN, embeddings, ensembles, PubMedBERT\\\hline
        dmiip\_fdu & d2v, tf-idf, SVM, KNN, LTR,  DeepMeSH, AttentionXML, BERT, PLT \\\hline
        Iria  & Luchene Index, multilabel k-NN, stem bigrams, ensembles, UIMA ConceptMapper\\\hline

        \end{tabular}
         \caption{Systems and approaches for Task 9a. Systems for which no description was available at the time of writing are omitted. }\label{tab:a_sys}
\end{table}

The team of Roche and Bogazici University participated in task 9a with four different systems (``\textit{bert\_dna}'' and ``\textit{pi\_dna}'' variations). In particular, their systems are based on the BERT framework with SentencePiece tokenization, and multiple binary classifiers. 
The rest of the teams build upon existing systems that had already competed in previous versions of the task. 
The National Library of Medicine (NLM) team competed with five different systems \cite{Rae2021clef}. To improve their previously developed CNN model \cite{rae2020automatic}, they utilized a pretrained
transformer model, PubMedBERT, which was fine-tuned to rank candidates  obtained from the CNN.
The Fudan University (``\textit{dmiip\_fdu}'') team also relied on their previous ``\textit{AttentionXML}''~\cite{You2018}, ``\textit{DeepMeSH}''~\cite{peng2016}, and ``\textit{BERTMeSH}'' models \cite{you2021bertmesh}. Differently from their previous version, they extended AttentionXML with BioBERT.
Finally, the team of Universidade de Vigo and Universidade da Coruña competed with two systems (``\textit{Iria}'') that followed the same approach used by the systems in previous versions of the task \cite{Ribadas2015}.

As in previous versions of the challenge, two systems developed by NLM to facilitate the annotation of articles by indexers in MedLine/PubMed, where available as baselines for the semantic indexing task. MTI \cite{morkBioasq2014} as enhanced in \cite{zavorin2016} and an extension based on features suggested by the winners of the first version of the task \cite{tsoumakasBioasq}.

\subsection{Task 9b}

This version of Task 9b was undertaken by 90 different systems in total, developed by 24 teams. In phase A, 9 teams participated, submitting results from 34 systems. In phase B, the numbers of participants and systems were 20 and 70 respectively. There were only three teams that engaged in both phases.
An overview of the technologies employed by the teams is provided in Table \ref{tab:b_sys} for the systems for which a description was available. Detailed descriptions for some of the systems are available at the proceedings of the workshop.

\begin{table}[!htb]
        \centering
        \begin{tabular}{M{0.23\linewidth}M{0.07\linewidth}M{0.7\linewidth}}\hline
        \textbf{Systems} & \textbf{Phase}& \textbf{Approach} \\ \hline
        bio-answerfinder & A, B &  
            Bio-AnswerFinder, ElasticSearch, Bio-ELECTRA, ELECTRA, 
            BioBERT, SQuAD, wRWMD
        \\\hline 
        RYGH & A & BM25, BioBERT, PubMedBERT, T5\\\hline
        bioinfo	 & A & BM25, ElasticSearch, distant learning, DeepRank, universal weighting passage mechanism (UPWM), BERT  \\\hline
        KU-DMIS & B & 
            BioBERT, NLI, MultiNLI, SQuAD, 
            BART, beam search, BERN, language\_check, sequence\_tagging
        \\\hline

        MQ & B & BERT, ROUGE \\\hline 
        Ir\_sys & B & BM25, T5, BERT, SpanBERT, XLNet, PubmedBERT, BART \\\hline
        CRJ & B & Proximal Policy Optimization (PPO), word2vec, BERT, Reinforcement Learning \\\hline
        LASIGE\_ULISBOA & B & BioBERT, transfer learning \\\hline
        UvA & B & encoder-decoder model,  seq2seq, BART \\\hline
        MDS\_UNCC & B & BioBERT \\\hline
        ALBERT & B & DistilBERT, ALBERT, SQuAD \\\hline
        UDEL-LAB & B & BioM-ALBERT, BioM-ELECTRA, SQuAD \\\hline
        MQU & B & BERT, summarization \\\hline
        NCU-IISR/AS-GIS & B & BioBERT, PubMedBERT, logistic-regression \\\hline
        \hline
        \end{tabular}
        \caption{Systems and approaches for Task9b. Systems for which no information was available at the time of writing are omitted.}\label{tab:b_sys}
\end{table}


The ``\textit{UCSD}'' team \cite{Ozyurt2021clef} participated in both phases of the task with two systems (``\textit{bio-answerfinder}''). Specifically, for phase A they relied on previously developed Bio-AnswerFinder system~\cite{ozyurt2020bio}, but instead of LSTM based keyword selection classifier, they used a Bio-ELECTRA++ model based keyword selection classifier together with the Bio-ELECTRA Mid based re-ranker \cite{ozyurt2020effectiveness}. This model was also used as an initial step for their systems in phase B, in order to re-rank candidate sentences. For factoid and list questions they fine-tuned a Bio-ELECTRA model using both SQuad and BioASQ training data. The  answer candidates are then scored considering classification probability, the top ranking of corresponding snippets and number of occurrences. Finally a normalization and filtering step is performed and, for list questions, an enrichment step based on coordinated phrase detection. For yes/no questions, they used a Bio-ELECTRA model based ternary yes/no/neutral classifier. The final decision is made by score voting. 
For summary questions, they follow two approaches. First, they employ hierarchical clustering, based on weighted relaxed word mover's distance (wRWMD) similarity~\cite{ozyurt2020bio} to group the top sentences, and select the sentence ranked highest by Bio-AnswerFinder to be concatenated to form the summary. Secondly, an abstractive summarization system based on the unified text-to-text transformer model t5 \cite{raffel2019exploring} is used.

In phase A, the team from the University of Aveiro participated with four distinct ``\textit{bioinfo}'' systems \cite{Almeida2021clef}. Relying on their previous model \cite{almeida2020bit}, they improved the computation flow and experimented with the transformer architecture. In the end, they developed two variants that used the passage mechanism from \cite{almeida2020bit} and the BERT model. The ``\textit{RYGH}'' team participated in phase A with five systems. They adopted a pipeline that utilized the BM25 along with several pre-trained models including BioBERT, PubMedBERT, PubMedBERT-FullText and T5.

In phase B, this year the ``\textit{KU-DMIS}'' team \cite{Yoon2021clef} participated in both exact and ideal answers.
Their systems are based on the transformers models and follow either a model-centric or a data-centric approach. The former, which is based on the sequence tagging approach \cite{yoon2021sequence}, is used for list questions while the latter, which relies on the characteristics of the training datasets and therefore data cleaning and sampling are important aspects of its architecture, is used for factoid questions. 
For yes/no questions, they utilized the BioBERT-large model, as a replacement of the previously used BioBERT-BASE model.  
For ideal questions, they followed the last year's strategy, where their BART model utilizes the predicted exact answer as a input for generating an ideal answer. 

There were four teams from the Macquarie University that participated in task 9b. The first team (``\textit{MQ}'') \cite{Molla2021clef} competed with five systems which are based on the use of BERT variants in a classification setting. The classification task takes as input the question, a sentence, and the sentence position, and the target labels are based on the ROUGE score of the sentence with respect to the ideal answer. 
The second team (``\textit{CRJ}'')  competed with three systems that followed the Proximal Policy Optimization (PPO) approach to Reinforcement Learning \cite{molla2020query}, and also utilized word2vec and BERT word embeddings.
The third team (ALBERT) \cite{Khanna2021clef} competed with four systems that were based on the transformer-based language models, DistilBERT and ALBERT. The pretrained models were fine-tuned first on the SQuAD dataset and then on the BioASQ dataset. Finally, the fourth team (``\textit{MQU}'') participated with five systems. Their systems utilized sentence transformers fine-tuned for passage retrieval, as well as abstractive summarizers trained on news media data.

The Fudan University team participated with four systems (``\textit{Ir\_sys}''). All systems utilized variants of the BERT framework. For yes/no questions they used BioBERT, while for factoid/list questions they combined SpanBERT, PubmedBERT and XLNet. 
For summary questions, they utilized both extractive and abstractive methods. For the latter, they performed conditional generation of answers by employing the BART model. 
The ``\textit{LASIGE\_ULISBOA}'' team \cite{Campos2021clef}, from the University of Lisboa, competed with four systems which are based on BioBERT . The models are fine-tuned on larger non-medical datasets prior to training on the task's datasets. The final decisions for the list questions are computed by applying a voting scheme, while a softmax is utilized for the remaining questions.  

The University of Delaware team \cite{Alrowili2021clef} participated with four systems (``\textit{UDEL-LAB}'') which are based on BioM-Transformets models \cite{alrowili-shanker-2021-biom}. In particular, they used both BioM-ALBERT and BioM-ELECTRA, and also applied transfer learning by fine tuning the models on MNLI and SQuAD datasets.
The ``\textit{NCU-IISR}'' team \cite{Zhang2021clef}, as in the previous version of the challenge,  participated in both parts of phase B, constructing various BERT-based models. In particular, they utilized BioBERT and PubMedBERT models to score candidate sentences. Then, as a second step a logistic regressor, trained on predicting the similarity between a question and each snippet sentence, re-ranks the sentences.

The ``\textit{Universiteit van Amsterdam}'' team submitted three systems (``\textit{UvA}'') that focused on ideal answers. They reformulated the task as a seq2seq language generation task in an encoder-decoder setting. All systems utilized variants of pre-trained language generation models. Specifically, they used BART and MT5 \cite{xue2020mt5}.

In this challenge too, the open source OAQA system proposed by \cite{yang2016learning} served as baseline for phase B exact answers. The system which achieved among the highest performances in previous versions of the challenge remains a strong baseline for the exact answer generation task. The system is developed based on the UIMA framework. ClearNLP is employed for question and snippet parsing. MetaMap, TmTool \cite{Wei2016}, C-Value and LingPipe \cite{baldwin2003lingpipe} are used for concept identification and UMLS Terminology Services (UTS) for concept retrieval. The final steps include identification of concept, document and snippet relevance based on classifier components and scoring and finally ranking techniques.

\subsection{Task MESINESP}
MESINESP track received greater interest from the public in this second edition. Out of 35 teams registered for CLEF Labs 2021, 7 teams from China, Chile, India, Spain, Portugal and Switzerland finally took part in the task. These teams provided a total of 25 systems for MESINESP-L, 20 for MESINESP-T and 20 for MESINESP-P. Like last year, the approaches were pretty similar to those of the English track, relying mainly on deep language models for text representation using BERT-based systems and extreme multilabel classification strategies.

Table \ref{tab:mesinesp_sys} describes the general methods used by the participants. Most of the teams used sophisticated systems such as AttentionXML, graph-based entity linking, or label encoding systems. But unlike the first edition, this year some teams have also tested models with more traditional technologies such as TF-IDF to evaluate their performance in the indexing of documents in Spanish, 

This year's baseline was an improved textual search system that searches the text for both DeCS descriptors and synonyms to assign codes to documents. This approach got an MiF of 0.2876 for scientific literatre, 0.1288 for clinical trials and 0.2992 for patents.

\begin{table}[!htb]
        \centering
        \begin{tabular}{M{0.3\linewidth}M{0.1\linewidth}M{0.5\linewidth}}\hline
        \textbf{System} & \textbf{Ref} & \textbf{Approach} \\ \hline
        Iria &  & k-NN, Luchene Index, lemmas, syntactic dependencies, NP, chunks, name entities, Sentence embedings\\\hline
        Fudan University & - & AttentionXML, Multilingual BERT, label-level attention \\\hline
        Roche & \cite{hybrid-pi-dna}  & SentencePiece, NER, BETO, multiple binary classifiers, synonym matching \\\hline
        Vicomtech & \cite{agarciap2021mesinesp} & BERT based classifier, label encoding\\\hline 
        LASIGE & \cite{ruas2021} & Graph-based entity linking, Personalized PageRank, semantic similarity-based filter, X-Transformer, Multilingual BERT \\\hline
        Universidad de Chile & - & TF-IDF, word embeddings, cosine similarity \\\hline
        \end{tabular}
         \caption{Systems and approaches for Task MESINESP 2021. Systems for which no description was available at the time of writing are omitted. }\label{tab:mesinesp_sys}
\end{table}

\subsection{Task Synergy}

In the first two versions of the new task Synergy, introduced this year, 15 teams participated submitting the results from 39 distinct systems. 
An overview of systems and approaches employed in this task is provided in Table \ref{tab:syn} for the systems for which a description was available. More detailed descriptions for some of the systems are available at the proceedings of the workshop.

\begin{table}[!htb]
        \centering
        \begin{tabular}{M{0.2\linewidth}M{0.8\linewidth}}\hline
        \textbf{System} & \textbf{Approach} \\ \hline
        RYGH  & BM25, BioBERT, SciBERT, ELECTRA, Text-to-Text Transfer Transformer (T5), Reciprocal Rank Fusion (RRF),Named Entity Recognition, BERT, SQuAD, SpanBERT \\\hline 
        bio-answerfinder & Bio-ELECTRA++, BERT, weighted relaxed word mover's distance (wRWMD), pyserini with MonoT5, SQuAD, GloVe \\\hline 
        AUEB &  BM25, Word2Vec, Graph-Node Embeddings, SciBERT, DL (JPDRMM) \\\hline 
        MQ & Word2Vec, BERT, LSTM, Reinforcement Learning (PPO)  \\\hline 
       bioinfo & BM25, ElasticSearch, distant learning, DeepRank, universal weighting passage mechanism (UPWM), BERT  \\\hline 
        NLM & BM25 model, T5, BART    \\\hline 
        pa-synergy & Lucene full-text search, BERT \\\hline
        \end{tabular}
         \caption{Systems and their approaches for Task Synergy. Systems for which no description was available at the time of writing are omitted. }\label{tab:syn}
\end{table}

The \textit{Fudan University} team, uses BM25 to fetch the top documents and then they use BioBERT, SciBERT, ELECTRA and T5 models to score the relevance of each document and query. 
Finally, the reciprocal rank fusion (RRF) is used to get the final document ranking results by integrating the previous results. Similarly, for snippet retrieval task, we use the same method with the focus in sentence. 
They also participate in all four types of questions. For the Yes/No type, they use the BERT encoder, a linear transformation layer and the sigmoid function to calculate the yes or no probability. 
For Factoid/List questions, they again employ BERT as the backbone and fine-tune the model with SQuAD. 
For Summary questions, they perform conditional generation of answers by adopting BART as the backbone of the model. 
As this is a collaborative task, they use experts' feedback data in two aspects: one is to expand query by Named Entity Recognition, and the other is to finetune the model by using feedback data.

The \textit{``MQ"} team \cite{Molla2021clef} focused on the question answering component of the task, section ideal answers using one of their systems that participated in BioASQ 8b~\cite{molla2020query} Phase B
For document retrieval, they used the top documents returned by the API provided by BioASQ.
For snippet retrieval, they re-ranked the document sentences based on tfidf-cosine similarity with the question or the sentence score predicted by their QA system. In run 4, they experimented with a variant of document retrieval based on an independent retrieval system, tuned with the BioASQ data.
What is more, they incorporated feedback from previous rounds to remove false negatives in documents and snippets and omit all documents and snippets that had been previously judged.

The \textit{``bio-answerfinder"} team \cite{Ozyurt2021clef} used the the Bio-AnswerFinder end-to-end QA system they had previously developed \cite{ozyurt2020bio}.
For exact answers and ideal answers they used re-ranked candidate sentences as input to the Synergy challenge subsystems. For factoid and list questions they used an answer span classifier fine-tuned ELECTRA\_Base~\cite{clark2020electra} using combined SQuAD v1.1 and BioASQ 8b training data. For list questions, answer candidates were enriched by coordinated phrase detection and processing.
For Yes/No questions, they used a binary classifier fine-tuned on ELECTRA Base using training data created/annotated from BioASQ 8b training set (ideal answers). 
For summary questions, they used the top 10 selected sentences to generate an answer summary. Hierarchical clustering using weighted relaxed word mover's distance (wRWMD) similarity was used to group sentences, with similarity threshold to maximize ROGUE-2 score.
They used the feedback provided to augment the training data used for the BERT [3] based reranker classifier used by Bio-AnswerFinder, after weighted relaxed word mover's distance (wRWMD) similarity based ranking and focus-word-based filtering. At each round, the BERT-Base based reranker was retrained with the cumulative Synergy expert feedback.

The \textit{``University of Aveiro''} team \cite{Almeida2021clefSynergy} built on their BioASQ Task 8b implementation~\cite{almeida2020bit} modifying it to fit the Synergy Task by adding methodology for the given feedback of each round. Their approach was to create 
a strong baseline using simple relevance feedback technique, using a tf-idf score they expanded the query and finally processed it using the BM25 algorithm.
This approach was adopted for questions having some feedback from previous rounds, for the new questions they used the BM25 algorithm along with reranking, similarly to the BioASQ Task 8b.
The \textit{``NLM"} team \cite{Sarrouti2021clef} first used the BM25 model to retrieve relevant articles and reranked them with the Text-to-Text Transfer Transformer (T5) relevance-based reranking model. For snippets, after splitting the relevant articles into sentences and chunks they used a re-ranking model based on T5 relevance. For ideal answers they used extractive and abstractive approaches. For the former they concatenated the top-n snippets, while for the later they finetuned their model using Bidirectional and Auto-Regressive Transformers (BART) on multiple biomedical datasets. 

The \textit{``AUEB"} team also built on their implementation from BioASQ Task 8b~\cite{pappas2020aueb} exploiting the feedback to filter out the material that was already assessed. They participated in all stages of the Synergy Task. They use mostly JPDRMM-based methods with ElasticSearch for document retrieval and  SEMantic Indexing for SEntence Retrieval (SEMISER) for snippet retrieval.
The \textit{``JetBrains"} team were based on their BioASQ Task 8b approach as well. In short, they used Lucene full-text search combined with BERT based reranker for document retrieval and BERT-based models for exact answers, without using the feedback provided by the experts.
The \textit{``MQU"} team used sentence vector similarities on the entire CORD-19 dataset, not considering the expert feedback either.

\section{Results}
\label{sec:results}

\subsection{Task 9a}
\begin{table*}[!htbp]
\centering
\begin{tabular}{M{0.3\linewidth}M{0.1\linewidth}M{0.1\linewidth}M{0.1\linewidth}M{0.1\linewidth}M{0.1\linewidth}M{0.1\linewidth}}\hline
\textbf{System} & \multicolumn{2}{c}{\textbf{Batch 1}} & \multicolumn{2}{c}{\textbf{Batch 2}} & \multicolumn{2}{c}{\textbf{Batch 3}} \\ \hline
& MiF & LCA-F & MiF & LCA-F & MiF & LCA-F \\ \cline{2-7}
dmiip\_fdu             & \textbf{1.25}        & \textbf{1.25}        & \textbf{1.375}       & \textbf{1.875}            & \textbf{2}           & \textbf{2.25}        \\
deepmesh\_dmiip\_fdu   & 2.25        & 2           & 3.375      & 2.75             & 3.25        & 2.75       \\
attention\_dmiip\_fdu  & 2.75                 & 3                    & 1.5                  & \textbf{1.875}            & 2.25                 & \textbf{2.25}        \\
deepmesh\_dmiip\_fdu\_ & 3.5                  & 3.5                  & 3                    & 2.375 & 2.75                 & 2.875                \\
NLM System 3           & 4                    & 4                    & 4.75                 & 4.75                      & 3                    & 2.875                \\
NLM System 1           & 5.25                 & 5.25                 & 6.5                  & 6.5                       & 7.5                  & 7.5                  \\
MTI First Line Index   & 6.75                 & 6.5                  & 7.75                 & 7.75                      & 8.5                  & 8.5                  \\
Default MTI            & 7.75                 & 7.5                  & 8.5                  & 8.5                       & 10                   & 9.5                  \\
NLM CNN                & 8.75                 & 8.75                 & 9.75                 & 9.75                      & 11.25                & 12.5                 \\
pi\_dna\_2             & -                    & -                    & 11.375               & 12    & 14.25                & 14.25                \\
pi\_dna                & -                    & -                    & 12.25                & 12.5                      & 9.25                 & 10.25                \\
bert\_dna              & -                    & -                    & 12.75                & 12.75                     & 12.375               & 11.5                 \\
iria-1                 & -                    & -                    & 14.875               & 15.125                    & 16.125               & 16.25                \\
iria-mix               & -                    & -                    & 15.125               & 14.875                    & 17.25                & 17.125               \\
DeepSys1               & -                    & -                    & 15.5                 & 15.25                     & -                    & -                    \\
NLM System 2           & -                    & -                    & -                    & -                         & 5.5                  & 5                    \\
NLM System 4           & -                    & -                    & -                    & -                         & 6.5                  & 6.5                  \\
pi\_dna\_3             & -                    & -                    & -                    & -                         & 12.75                & 13                   \\
DeepSys2               & -                    & -                    & -                    & -                         & 15.25                & 15.25                \\
\hline
\end{tabular}
\caption{Average system ranks across the batches of the task 9a. A hyphenation symbol (-) is used whenever the system participated in fewer than 4 test sets in the batch. Systems participating in fewer than 4 test sets in all three batches are omitted.}\label{tab:a_res}
\end{table*}

In Task 9a, each of the three batches were independently evaluated as presented in Table~\ref{tab:a_res}.
As in previous versions of the task, standard evaluation measures \cite{balikas13} were used for measuring the classification performance of the systems, both flat and hierarchical. In particular, the official measures  used to identify the winners for each batch were the micro F-measure (MiF) and the Lowest Common Ancestor F-measure (LCA-F) \cite{kosmopoulos2015evaluation}.
As suggested by Demšar \cite{Demsar06}, the appropriate way to compare multiple classification systems over multiple datasets is based on their average rank across all the datasets. 

\begin{figure*}[!htb]
\centerline{\includegraphics[width=1\textwidth]{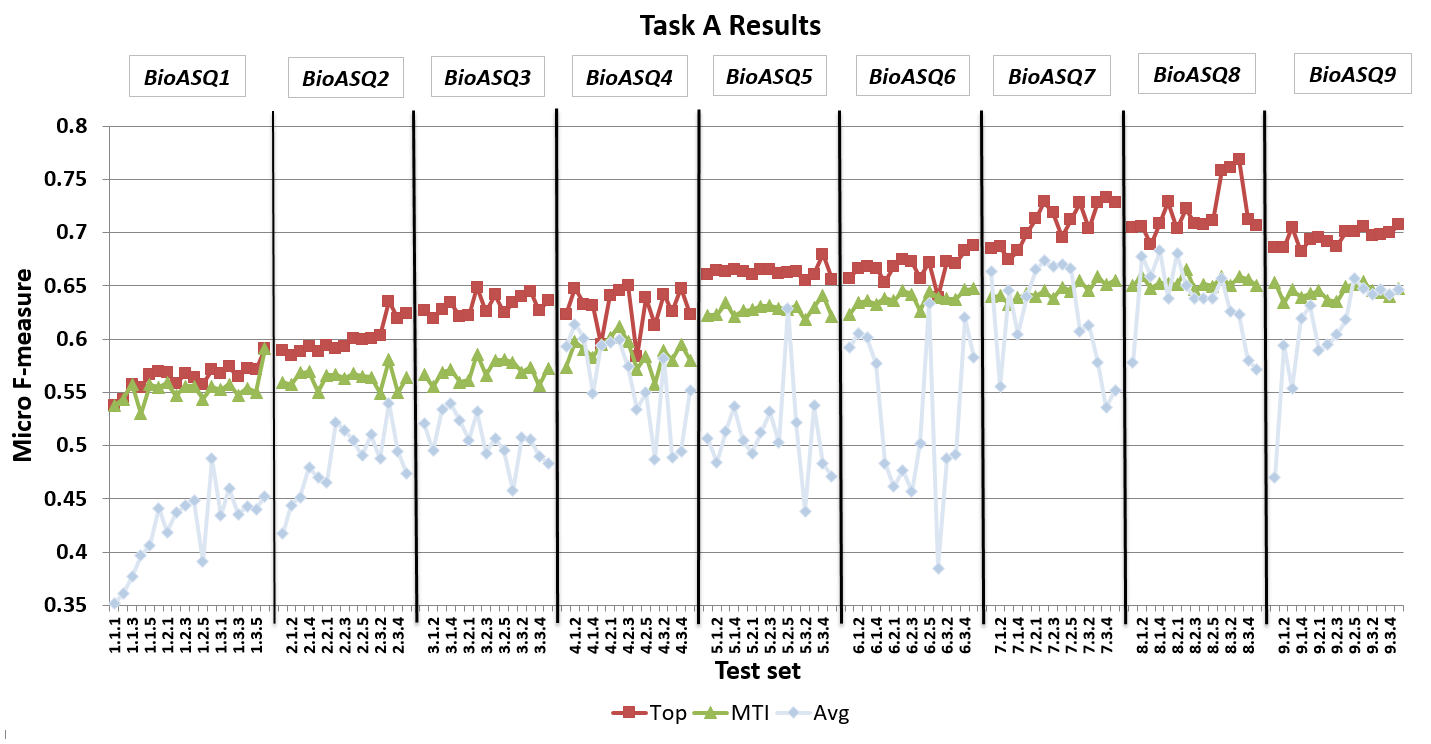}}
\caption{The micro f-measure (MiF) achieved by systems across different years of the BioASQ challenge. For each test set the MiF score is presented for the best performing system (Top) and the MTI, as well as the average micro f-measure of all the participating systems (Avg). }\label{fig:mif}
\end{figure*}

In this task, the system with the best performance in a test set gets rank 1.0 for this test set, the second best rank 2.0 and so on. 
In case two or more systems tie, they all receive the average rank.
Based on the rules of the challenge, the average rank of each system for a batch is the average of the four best ranks of the system in the five test sets of the batch.
The average rank of each system, based on both the flat MiF and the hierarchical LCA-F scores, for the three batches of the task are presented in Table~\ref{tab:a_res}.

The results of Task 9a reveal that several participating systems manage to outperform the strong baselines in all test batches and considering either the flat or the hierarchical measures. Namely, the ``\textit{dmiip\_fdu}'' systems from the Fudan University team achieve the best performance and the ``NLM'' systems the second best in all three batches of the task. More detailed results can be found in the online results page\footnote{\footnotesize \url{http://participants-area.bioasq.org/results/9a/}}. 
Figure {\ref{fig:mif}} presents the improvement of the MiF scores achieved by both the MTI baseline and the top performing participant systems through the nine years of the BioASQ challenge.

\subsection{Task 9b}
\textbf{Phase A}: 
The evaluation of phase A in Task 9b is based on the Mean Average Precision (MAP) measure for each of the three types of annotations, namely documents, concepts and RDF triples. 
For snippets, where several distinct snippets may overlap with the same golden snippet, interpreting the MAP, which is based on the number of relevant elements, is more complicated. 
Therefore, this year, the F-measure is used for the official ranking of the systems in snippet retrieval, which is calculated based on character overlaps\footnote{\url{http://participants-area.bioasq.org/Tasks/b/eval\_meas\_2021/}}.

As in BioASQ8, a modified version of Average Precision (AP) is adopted. 
In brief, since BioASQ3, the participant systems are allowed to return up to 10 relevant items (e.g. documents), and the calculation of AP was modified to reflect this change. However, some questions with fewer than 10 golden relevant items have been observed in the last years, resulting to relatively small AP values even for submissions with all the golden elements. Therefore, the AP calculation was modified to consider both the limit of 10 elements and the actual number of golden elements \cite{nentidis2020overview}.

\begin{table*}[!htbp]
\centering
\begin{tabular}{M{0.3\linewidth}M{0.13\linewidth}M{0.13\linewidth}M{0.13\linewidth}M{0.13\linewidth}M{0.12\linewidth}}\hline
\textbf{System} & \textbf{Mean Precision} & \textbf{Mean Recall} & \textbf{Mean F-measure} & \textbf{MAP} & \textbf{GMAP}  \\ \hline
bioinfo-2            & 0.1280               & 0.5213               & 0.1873               & \textbf{0.4236}               & 0.0125               \\
pa-5                 & \textbf{0.2421}               & 0.5132               & \textbf{0.2902}               & 0.4192               & 0.0128               \\
RYGH-4               & 0.1170               & 0.5118               & 0.1733               & 0.4179               & 0.0123               \\
RYGH-3               & 0.1150               & 0.5120               & 0.1726               & 0.4174               & \textbf{0.0132}               \\
RYGH                 & 0.1140               & 0.5083               & 0.1701               & 0.4166               & 0.0120               \\
RYGH-5               & 0.1160               & 0.5118               & 0.1733               & 0.4109               & 0.0122               \\
bioinfo-3            & 0.1270               & \textbf{0.5280}               & 0.1865               & 0.4042               & 0.0131               \\
bioinfo-4            & 0.1270               & \textbf{0.5280}               & 0.1865               & 0.4042               & 0.0131               \\
RYGH-1               & 0.1160               & 0.5110               & 0.1725               & 0.4027               & 0.0111               \\
pa-1                 & 0.1410               & 0.4773               & 0.1930               & 0.3893               & 0.0092               \\
\hline
\end{tabular}
\caption{Preliminary results for document retrieval in batch 4 of phase A of Task 9b. 
Only the top-10 systems are presented, based on MAP.
}\label{tab:bA_res_doc}

\centering
\begin{tabular}{M{0.3\linewidth}M{0.14\linewidth}M{0.12\linewidth}M{0.14\linewidth}M{0.12\linewidth}M{0.12\linewidth}}\hline
\textbf{System} & \textbf{Mean Precision} & \textbf{Mean Recall} & \textbf{Mean F-measure} & \textbf{MAP} & \textbf{GMAP}  \\ \hline
pa-5                 & \textbf{0.1932}               & 0.3147               & \textbf{0.2061}               & \textbf{0.9696}               & 0.0026               \\
RYGH-4               & 0.1416               & \textbf{0.3337}               & 0.1764               & 0.4561               & \textbf{0.0068}               \\
RYGH-1               & 0.1369               & 0.3334               & 0.1737               & 0.4697               & 0.0062               \\
pa-1                 & 0.1567               & 0.2777               & 0.1733               & 0.8515               & 0.0018               \\
RYGH-5               & 0.1397               & 0.3257               & 0.1733               & 0.4492               & 0.0064               \\
pa-2                 & 0.1563               & 0.2745               & 0.1722               & 0.8372               & 0.0015               \\
RYGH                 & 0.1382               & 0.3271               & 0.1722               & 0.4595               & 0.0059               \\
RYGH-3               & 0.1382               & 0.3234               & 0.1721               & 0.4490               & 0.0064               \\
pa-3                 & 0.1567               & 0.2686               & 0.1718               & 0.8437               & 0.0012               \\
pa-4                 & 0.1567               & 0.2686               & 0.1718               & 0.8437               & 0.0012               \\

        \hline
        \end{tabular}
        \caption{Preliminary results for snippet retrieval in batch 4 of phase A of Task 9b. Only the top-10 systems are presented, based on F-measure.
        }\label{tab:bA_res_sni}
\centering
\begin{tabular}
{M{0.205\linewidth}M{0.0852\linewidth}M{0.0852\linewidth}M{0.105\linewidth}M{0.11\linewidth}M{0.0852\linewidth}M{0.0852\linewidth}M{0.0852\linewidth}M{0.0852\linewidth}}
\hline

\textbf{System} & \multicolumn{2}{c}{\textbf{Yes/No}} & \multicolumn{3}{c}{\textbf{Factoid}} & \multicolumn{2}{c}{\textbf{List}} \\ 
\hline
& F1 & Acc. & Str. Acc. & Len. Acc. & MRR & Prec. & Rec. & F1 \\ \cline{2-9}    
KU-DMIS-1             & \textbf{0.9480} & \textbf{0.9600} & 0.5000 & 0.6071 & 0.5310 & 0.6454 & \textbf{0.8202} & \textbf{0.7061} \\
Ir\_sys1              & \textbf{0.9480} & \textbf{0.9600} & \textbf{0.6429} & \textbf{0.7857} & \textbf{0.6929} & 0.5929 & 0.7675 & 0.6312 \\
KU-DMIS-5             & 0.9008 & 0.9200 & 0.5000 & 0.7143 & 0.5726 & 0.6245 & 0.7377 & 0.6470 \\
KU-DMIS-2             & 0.8904 & 0.9200 & 0.5000 & 0.6786 & 0.5589 & 0.5568 & 0.7465 & 0.6001 \\
KU-DMIS-3             & 0.8904 & 0.9200 & 0.5000 & 0.6429 & 0.5429 & 0.5991 & 0.7860 & 0.6430 \\
KU-DMIS-4             & 0.8904 & 0.9200 & 0.4286 & 0.6786 & 0.5101 & 0.5521 & 0.7149 & 0.5802 \\
NCU-IISR...1     & 0.8441 & 0.8800 & 0.3571 & 0.6071 & 0.4232 & 0.5263 & 0.3991 & 0.4261 \\
NCU-IISR...2     & 0.8441 & 0.8800 & 0.3571 & 0.6071 & 0.4232 & 0.5263 & 0.3991 & 0.4261 \\
NCU-IISR...3     & 0.8441 & 0.8800 & 0.3571 & 0.6071 & 0.4232 & 0.5263 & 0.3991 & 0.4261 \\
Ir\_sys2              & 0.8252 & 0.8800 & 0.6071 & 0.7500 & 0.6464 & 0.6027 & 0.6614 & 0.5780 \\
BioASQ\_Baseline      & 0.3506 & 0.3600 & 0.1429 & 0.3571 & 0.2077 & 0.1767 & 0.3202 & 0.1857 \\

\hline
\end{tabular}
\caption{Results for batch 4 for exact answers in phase B of Task 9b.
Only the top-10 systems based on Yes/No F1 and the BioASQ Baseline are presented.
\label{tab:bB_res}}
\end{table*}

Some indicative preliminary results from batch 4 are presented in Tables \ref{tab:bA_res_doc} and \ref{tab:bA_res_sni} for document and snippet retrieval. The full results are available in the online results page of Task 9b, phase A\footnote{\footnotesize \url{http://participants-area.bioasq.org/results/9b/phaseA/}}. The results presented here are preliminary, as the final results for the task 9b will be available after the manual assessment of the system responses by the BioASQ team of biomedical experts. 

\textbf{Phase B}: 
In phase B of task 9b, both exact and ideal answers are expected by the participating systems. 
For the sub-task of ideal answer generation, the BioASQ experts assign manual scores to each answer submitted by the participating systems during the assessment of system responses~\cite{balikas13}. Then these scores are used for the official ranking of the systems. 
Regarding exact answers, the participating systems are ranked based on their average ranking in the three question types where exact answers are required. Summary questions are not considered as no exact answers are submitted for them.
For yes/no questions, the systems are ranked based on the F1-measure, macro-averaged over the class of no and yes. 
For factoid questions, the ranking is based on mean reciprocal rank (MRR) and for list questions on mean F1-measure.
Indicative preliminary results for exact answers from the fourth batch of Task 9b are presented in Table~\ref{tab:bB_res}. The full results of phase B of Task 9b are available online\footnote{\footnotesize \url{http://participants-area.bioasq.org/results/9b/phaseB/}}. These results are preliminary, as the final results for Task 9b will be available after the manual assessment of the system responses by the BioASQ team of biomedical experts.

\begin{figure*}[!htbp]
\centerline{\includegraphics[width=1\textwidth]{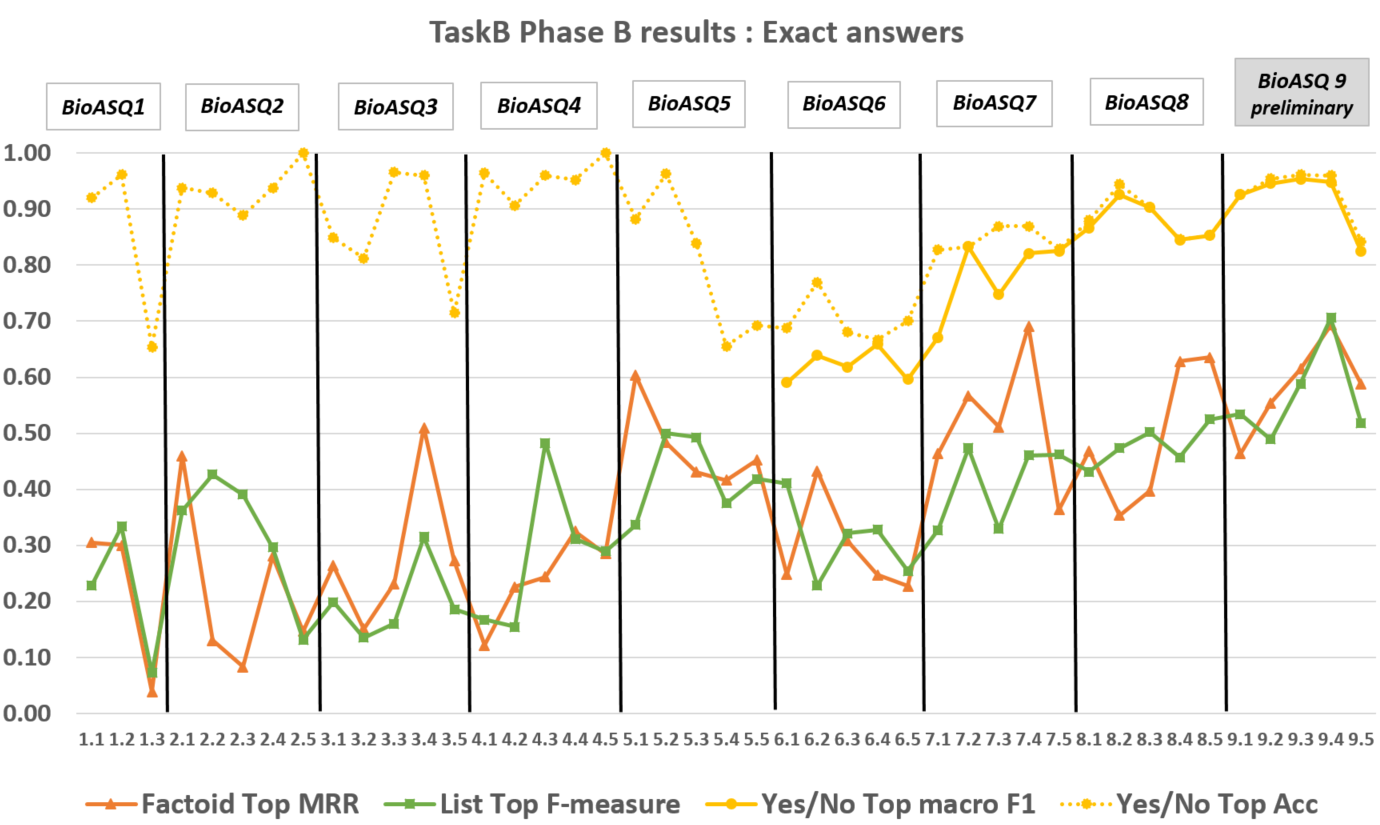}}
\caption{
The official evaluation scores of the best performing systems in Task B, 
Phase B, exact answer generation, across the nine years of the BioASQ
challenge. Since BioASQ6 the official measure for Yes/No questions is the
macro-averaged F1 score (macro F1), but accuracy (Acc) is also presented as the former official measure.
}\label{fig:Exact}
\end{figure*}

The top performance of the participating systems in exact answer generation for each type of question during the nine years of BioASQ is presented in Figure {\ref{fig:Exact}}.
These results reveal that the participating systems keep improving in all types of questions.
In batch 4, for instance, presented in Table \ref{tab:bB_res}, in yes/no questions most systems manage to outperform by far the strong baseline, which is based on a version of the OAQA system that achieved top performance in previous years.
Improvements are also observed in the preliminary results 
For list and factoid questions, some improvements are also observed in the preliminary results compared to the previous years, but there is still more room for improvement.

\newpage
\subsection{Task MESINESP}

The performance of participating teams this year is higher than last year. There has been an increase in f-score of 0.06 for scientific literature, and the state of the art of clinical trials and patents semantic indexing with DeCS has been established in 0.3640 and 0.4514.

As shown in Table \ref{tab:mesinesp2_results}, once again, the top performer this year was the BertDeCS system developed by Fudan University. Their system was based on an AttentionXML architecture with an  Multilingual BERT encoding layer that was trained with MEDLINE articles and then fine-tuned with MESINESP corpora. This architecture obtained the best MiF score performance in scientific literature, clinical trials and patents. However, the best code prediction accuracy was achieved by Roche's ``\textit{pi\_dna}" system. Comparing the performance of the models with the baseline, it is noteworthy that only 7 of the models implemented for patents have been able to outperform the look-up system, highlighting the good performance of \textit{iria-2}.

\begin{table}[!htp]
\centering
\resizebox{\textwidth}{!}{%
\begin{tabular}{ccccc}
\multicolumn{1}{l}{} & \multicolumn{1}{l}{} & \textbf{MESINESP-L} & \textbf{MESINESP-T} & \textbf{MESINESP-P} \\ \hline
\textbf{Team}                           & \textbf{System}        & \textbf{MiF}             & \textbf{MiF}             & \textbf{MiF}             \\ \hline
\multirow{5}{*}{Fudan University}       & BERTDeCS-CooMatInfer   & 0.4505                   & 0.1095                   & 0.4489                   \\
                                        & BERTDeCS version 2     & 0.4798                   & \textit{\textbf{0.3640}} & \textit{\textbf{0.4514}} \\
                                        & BERTDeCS version 3     &  0.4808            &  0.3630             & 0.4480                   \\
                                        & BERTDeCS version 4     & \textit{\textbf{0.4837}} & 0.3563                   &  0.4514             \\
                                        & bertmesh-1             & 0.4808                   & 0.3600                   & 0.4489                   \\ \hline
\multirow{5}{*}{Roche}                  & bert\_dna              & 0.3989                   & 0.2710                   & 0.2479                   \\
                                        & pi\_dna                & \textit{0.4225}          & \textit{0.2781}          & \textit{0.3628}          \\
                                        & pi\_dna\_2             & 0.3978                   & 0.2680                   & -                        \\
                                        & pi\_dna\_3             & 0.4027                   & -                        & -                        \\
                                        & bert\_dna\_2           & 0.3962                   & 0.2383                   & 0.2479                   \\ \hline
\multirow{5}{*}{Universidade de Lisboa} & LASIGE\_BioTM\_1       & \textit{0.2007}          & -                        & -                        \\
                                        & LASIGE\_BioTM\_2       & 0.1886                   & -                        & -                        \\
                                        & clinical\_trials\_1.0  & -                        & 0.0679                   & -                        \\
                                        & clinical\_trials\_0.25 & -                        & \textit{0.0686}          & -                        \\
                                        & patents\_1.0           & -                        & -                        & \textit{0.0314}          \\ \hline
\multirow{5}{*}{Vicomtech}              & Classifier             & \textit{0.3825}          & 0.2485                   & 0.1968                   \\
                                        & CSSClassifier025       & 0.3823                   & \textit{0.2819}          & 0.2834                   \\
                                        & CSSClassifier035       & 0.3801                   & 0.2810                   & 0.2651                   \\
                                        & LabelGlosses01         & 0.3704                   & 0.2807                   & 0.2908                   \\
                                        & LabelGlosses02         & 0.3746                   & -                        & \textit{0.2921}          \\ \hline
\multirow{5}{*}{Uni Vigo, Uni. Coruña}  & iria-1                 & 0.3406                   & \textit{0.2454}          & 0.1871                   \\
                                        & iria-2                 & 0.3389                   & -                        & \textit{0.3203}          \\
                                        & iria-3                 & 0.2537                   & 0.1562                   & 0.0793                   \\
                                        & iria-4                 & 0.3656                   & 0.2003                   & 0.2169                   \\
                                        & iria-mix               & \textit{0.3725}          & 0.2003                   & 0.2542                   \\ \hline
Universidad de Chile                    & tf-idf-model           & \textit{0.1335}          & -                        & -                        \\ \hline
\multirow{4}{*}{YMCA University}        & AnujTagging            & \textit{0.0631}          & -                        & -                        \\
                                        & Anuj\_ml               & -                        & \textit{0.0019}          & -                        \\
                                        & Anuj\_NLP              & 0.0035                   & -                        & -                        \\
                                        & Anuj\_Ensemble         & -                        & -                        & \textit{0.0389}          \\ \hline
\multicolumn{2}{c}{\textbf{Baseline}}                            & 0.2876                   & 0.1288                   & 0.2992                   \\ \hline
\end{tabular}%
}
\caption{Results of models }
\label{tab:mesinesp2_results}
\end{table}

The results of the task show a drop in performance compared to the English task despite teams using similar technologies. This drop in performance could be associated with a lower number of training documents and inconsistencies in the manual indexing of these documents because they come from two different bibliographic sources \cite{rodriguez2020overview}. Alternatively, this could also be explained by the delay in updating deprecated DeCS codes from the historical database. DeCS add and remove new terms twice a year, and the lack of temporal alignment in the update process could lead to inconsistencies between training and test data and decrease overall performance.

Regarding MESINESP-T track, there is no similar task in English to compare the results. The performance of the models is systematically lower than those generated for scientific literature. Because participants reported that they reused the models trained with scientific literature, incorporating the development set to make their predictions, a low quality Gold Standard cannot be associated with the drop in performance. However, given that the length of clinical trial documents is much longer than article abstracts, and that most systems use BERT models with an input size limit of 512 tokens, it is possible that a significant part of the documents will not be processed by the models and relevant information will be lost for indexing.

The patents subtrack presented a major challenge for the participants as they did not have a large training and development dataset. Since the statistics between the MESINESP-T and MESINESP-P corpora were similar, the participants solved the lack of data using the same models generated for scientific literature. The resulting models were promising, and the performance of some of the systems, such as Fudan, Roche and Iria, remained at the same level as scientific literature track.

On the other hand, although the performance of the models is lower than that of the English task, we used the participants' results to see whether the manual annotation process could be improved. To this end, a module for indexing assistance was developed in the ASIT tool, and a set of pre-annotated documents with the predictions of the best-performing team was provided to our expert indexers. After tracking annotation times, we observed that this type of system could improve annotation times by \textbf{up to 60\%}\cite{luis2020overview}.

\subsection{Synergy Task}

In task Synergy the participating systems were expected to retrieve documents and snippets, as in phase A of task B, and, at the same time, provide answers for some of these questions, as in phase B of task B. 
In contrast to task B, it is possible that no answer exists for some questions. Therefore only some of the questions provided in each test set, that were indicated to have enough relevant material gathered from previous rounds, require the submission of exact and ideal answers.
Also in contrast to task B, during the first round no golden documents and snippets were given, while on the rest of the rounds a separate file with feedback from the experts, based on the previously submitted responses, was provided.

The feedback concept was introduced in this task to further assist the collaboration between the systems and the BioASQ team of biomedical experts. The feedback includes the already judged documentation and answers along with their evaluated relevancy to the question. The documents and snippets included in the feedback are not considered valid for submission in the following rounds, and even if accidentally submitted, they will not be taken into account for the evaluation of that round. The evaluation measures for the retrieval of documents and snippets are the MAP and F-measure respectively, as in phase A of task B.

Regarding the ideal answers, the systems are ranked according to manual scores assigned to them by the BioASQ experts during the assessment of systems responses as in phase B of task B~\cite{balikas13}. 
For the exact answers, which are required for all questions except the summary ones, the measure considered for ranking the participating systems depends on the question type. 
For the yes/no questions, the systems were ranked according to the macro-averaged F1-measure on prediction of no and yes answer. 
For factoid questions, the ranking was based on mean reciprocal rank (MRR) and for list questions on mean F1-measure.

Some indicative results for the first round of Synergy Task, version 1, are presented for document retrieval in Table~\ref{tab:synergy1-res}.
The full results of Synergy Task are available online\footnote{\footnotesize \url{http://participants-area.bioasq.org/results/synergy/}}. 
As regards the extraction of exact answers, despite the moderate scores in list and factoid questions the experts found useful the submissions of the participants, as most of them (more than 70\%) stated they would be interested in using a tool following the BioASQ Synergy process to identify interesting material and answers for their research.

\begin{table}
    \centering
    \begin{tabular}{M{0.3\linewidth}M{0.14\linewidth}M{0.12\linewidth}M{0.14\linewidth}M{0.12\linewidth}M{0.12\linewidth}}
    \hline
        \textbf{System} & \textbf{Mean precision} & \textbf{Mean Recall} & \textbf{Mean F-Measure} & \textbf{MAP} & \textbf{GMAP} \\ \hline
        RYGH-5 & \textbf{0.4963} &\textbf{ 0.3795 }&\textbf{ 0.3457} & \textbf{0.3375} &\textbf{ 0.0829 }\\ 
        RYGH-3 & 0.4948 & 0.354 & 0.3454 & 0.3363 & 0.0418 \\ 
        RYGH-1 & 0.4892 & 0.3523 & 0.3358 & 0.3248 & 0.0471 \\ 
        RYGH-4 & 0.4799 & 0.3603 & 0.328 & 0.3236 & 0.0598 \\ 
        NLM-1 & 0.4773 & 0.3251 & 0.3383 & 0.2946 & 0.0459 \\ 
        NLM-2 & 0.4773 & 0.3251 & 0.3383 & 0.2946 & 0.0459 \\ 
        NLM-3 & 0.4438 & 0.331 & 0.3078 & 0.2735 & 0.0635 \\ 
        NLM-4 & 0.4438 & 0.331 & 0.3078 & 0.2735 & 0.0635 \\ 
        RYGH & 0.4225 & 0.3308 & 0.3016 & 0.3008 & 0.0281 \\ 
        bio-answerfinder & 0.4105 & 0.216 & 0.2372 & 0.1935 & 0.014 \\ 
        \hline 
    \end{tabular}
    \caption{Results for document retrieval in round 1 of the first version of Synergy task. Only the top-10 systems are presented.}\label{tab:synergy1-res}
\end{table}

\section{Conclusions}
\label{sec:conclusion}

An overview of the ninth BioASQ challenge is provided in this paper. This year, the challenge consisted of four tasks: The two tasks on biomedical semantic indexing and question answering in English, already established through the previous eight years of the challenge, the second version of the MESINESP task on semantic indexing of medical content in Spanish, and the new task Synergy on question answering for COVID-19. 

In the second version of the MESINESP task we introduced two new challenging sub-tracks, beyond the one on medical literature. Namely, on patents and clinical trials in Spanish. Due to the lack of big datasets in these new tracks, the participants were pushed to experiment with transferring knowledge and models from the literature track, highlighting the importance of adequate resources for the development of systems to effectively help biomedical experts dealing with non-English resources. 

The introduction of the Synergy Task, in an effort to enable a dialogue between the participating systems with biomedical experts revealed that state-of-the-art systems, despite they still have room for improvement, can be a useful tool for biomedical experts that need specialized information in the context of the developing problem of the COVID-19 pandemic.

The overall shift of participant systems towards deep neural approaches observed during the last years, is even more apparent this year. 
State-of-the-art methodologies have been successfully adapted to biomedical question answering and novel ideas have been explored leading to improved results, particularly for exact answer generation this year.
Most of the teams developed systems based on neural embeddings, such as BERT, SciBERT, and BioBERT models, for all tasks of the challenge.
In the QA tasks in particular, different teams attempted transferring knowledge from general domain QA datasets, notably SQuAD, or from other NLP tasks such as NER and NLI.

Overall, the top preforming systems were able to advance over the state of the art, outperforming the strong baselines on the challenging tasks offered in BioASQ, as in previous versions of the challenge.
Therefore, BioASQ keeps pushing the research frontier in biomedical semantic indexing and question answering, extending beyond the English language, through MESINESP, and beyond the already established models for the shared tasks, by introducing Synergy. 
The future plans for the challenge include the extension of the benchmark data for question answering though a community-driven process, as well as extending the Synergy task into other developing problems beyond COVID-19.

\section{Acknowledgments}
Google was a proud sponsor of the BioASQ Challenge in 2020. The ninth edition of BioASQ is also sponsored by the Atypon Systems inc. 
BioASQ is grateful to NLM for providing the baselines for task 9a and to the CMU team for providing the baselines for task 9b.
The MESINESP task is sponsored by the Spanish Plan for advancement of Language Technologies (Plan TL) and the Secretaría de Estado para el Avance Digital (SEAD).
BioASQ is also grateful to LILACS, SCIELO and Biblioteca virtual en salud and Instituto de salud Carlos III for providing data for the BioASQ MESINESP task.
%
%
%
\bibliographystyle{splncs04}
\bibliography{BioASQ8.bib}

\begin{thebibliography}{10}
\providecommand{\url}[1]{\texttt{#1}}
\providecommand{\urlprefix}{URL }
\providecommand{\doi}[1]{https://doi.org/#1}

\bibitem{You2018}
{}attentionxml: Label tree-based attention-aware deep model for
  high-performance extreme multi-label text classification

\bibitem{Yoon2021clef}
{}ku-dmis at bioasq 9: Data-centric and model-centric approaches for biomedical
  questionanswering

\bibitem{almeida2020bit}
Almeida, T., Matos, S.: Bit. ua at bioasq 8: Lightweight neural document
  ranking with zero-shot snippet retrieval. In: CLEF (Working Notes) (2020)

\bibitem{Almeida2021clefSynergy}
Almeida, T., Matos, S.: Bioasq synergy: A strong and simple baseline rooted in
  relevance feedback. In: CLEF (Working Notes) (2021)

\bibitem{Almeida2021clef}
Almeida, T., Matos, S.: Universal passage weighting mecanism (upwm) in bioasq
  9b. In: CLEF (Working Notes) (2021)

\bibitem{Alrowili2021clef}
Alrowili, S., Shanker, K.: Large biomedical question answering models with
  albert and electra. In: CLEF (Working Notes) (2021)

\bibitem{alrowili-shanker-2021-biom}
Alrowili, S., Shanker, V.: {B}io{M}-transformers: Building large biomedical
  language models with {BERT}, {ALBERT} and {ELECTRA}. In: Proceedings of the
  20th Workshop on Biomedical Language Processing. pp. 221--227. Association
  for Computational Linguistics, Online (Jun 2021),
  \url{https://www.aclweb.org/anthology/2021.bionlp-1.24}

\bibitem{baldwin2003lingpipe}
Baldwin, B., Carpenter, B.: Lingpipe. Available from World Wide Web:
  http://alias-i. com/lingpipe  (2003)

\bibitem{balikas13}
Balikas, G., Partalas, I., Kosmopoulos, A., Petridis, S., Malakasiotis, P.,
  Pavlopoulos, I., Androutsopoulos, I., Baskiotis, N., Gaussier, E., Artieres,
  T., Gallinari, P.: Evaluation framework specifications. Project
  deliverable~D4.1, UPMC (05/2013 2013)

\bibitem{Campos2021clef}
Campos, M., Couto, F.: Post-processing biobert and using voting methods for
  biomedical question answering. In: CLEF (Working Notes) (2021)

\bibitem{clark2020electra}
Clark, K., Luong, M.T., Le, Q.V., Manning, C.D.: Electra: Pre-training text
  encoders as discriminators rather than generators. arXiv preprint
  arXiv:2003.10555  (2020)

\bibitem{Demsar06}
Demsar, J.: Statistical comparisons of classifiers over multiple data sets.
  Journal of Machine Learning Research  \textbf{7},  1--30 (2006)

\bibitem{agarciap2021mesinesp}
Garc{\'i}a-Pablos, A., Perez, N., Cuadros, M.: {Vicomtech at MESINESP2:
  BERT-based Multi-label Classification Models for Biomedical Text Indexing}
  (2021)

\bibitem{luis2020overview}
Gasco, L., Nentidis, A., Krithara, A., Estrada-Zavala, D., , Murasaki, R.T.,
  Primo-Peña, E., Bojo-Canales, C., Paliouras, G., Krallinger, M.: {Overview
  of BioASQ 2021-MESINESP track. Evaluation of advance hierarchical
  classification techniques for scientific literature, patents and clinical
  trials.}  (2021)

\bibitem{hybrid-pi-dna}
Huang, Y., Buse, G., Abdullatif, K., Ozgur, A., Ozkirimli, E.: Pidna at bioasq
  mesinesp: Hybrid semanticindexing for biomedical articles in spanish  (2021)

\bibitem{Khanna2021clef}
Khanna, U., Molla, D.: Transformer-based language models for factoid question
  answering at bioasq9b. In: CLEF (Working Notes) (2021)

\bibitem{kosmopoulos2015evaluation}
Kosmopoulos, A., Partalas, I., Gaussier, E., Paliouras, G., Androutsopoulos,
  I.: Evaluation measures for hierarchical classification: a unified view and
  novel approaches. Data Mining and Knowledge Discovery  \textbf{29}(3),
  820--865 (2015)

\bibitem{krallinger2015overview}
Krallinger, M., Rabal, O., Louren{\c{c}}o, A., Perez, M.P., Rodriguez, G.P.,
  Vazquez, M., Leitner, F., Oyarzabal, J., Valencia, A.: Overview of the
  chemdner patents task. In: Proceedings of the fifth BioCreative challenge
  evaluation workshop. pp. 63--75 (2015)

\bibitem{miranda2020named}
Miranda-Escalada, A., Farr{\'e}, E., Krallinger, M.: Named entity recognition,
  concept normalization and clinical coding: Overview of the cantemist track
  for cancer text mining in spanish, corpus, guidelines, methods and results.
  In: Proceedings of the Iberian Languages Evaluation Forum (IberLEF 2020),
  CEUR Workshop Proceedings (2020)

\bibitem{miranda2021profner}
Miranda-Escalada, A., Farr{\'e}-Maduell, E., Lima-L{\'o}pez, S., Gasc{\'o}, L.,
  Briva-Iglesias, V., Ag{\"u}ero-Torales, M., Krallinger, M.: The profner
  shared task on automatic recognition of occupation mentions in social media:
  systems, evaluation, guidelines, embeddings and corpora. In: Proceedings of
  the Sixth Social Media Mining for Health (\# SMM4H) Workshop and Shared Task.
  pp. 13--20 (2021)

\bibitem{miranda2020overview}
Miranda-Escalada, A., Gonzalez-Agirre, A., Armengol-Estap{\'e}, J., Krallinger,
  M.: Overview of automatic clinical coding: annotations, guidelines, and
  solutions for non-english clinical cases at codiesp track of clef ehealth
  2020. In: Working Notes of Conference and Labs of the Evaluation (CLEF)
  Forum. CEUR Workshop Proceedings (2020)

\bibitem{molla2020query}
Molla, D., Jones, C., Nguyen, V.: Query focused multi-document summarisation of
  biomedical texts. arXiv preprint arXiv:2008.11986  (2020)

\bibitem{Molla2021clef}
Molla, D., Khanna, U., Galat, D., Nguyen, V., Rybinski, M.: Query-focused
  extractive summarisation for finding ideal answers to biomedical and covid-19
  questions. In: CLEF (Working Notes) (2021)

\bibitem{morkBioasq2014}
Mork, J.G., Demner-Fushman, D., Schmidt, S.C., Aronson, A.R.: Recent
  enhancements to the nlm medical text indexer. In: Proceedings of Question
  Answering Lab at CLEF (2014)

\bibitem{nentidis2020overview}
Nentidis, A., Krithara, A., Bougiatiotis, K., Krallinger, M.,
  Rodriguez-Penagos, C., Villegas, M., Paliouras, G.: Overview of bioasq 2020:
  The eighth bioasq challenge on large-scale biomedical semantic indexing and
  question answering. In: International Conference of the Cross-Language
  Evaluation Forum for European Languages. pp. 194--214. Springer (2020)

\bibitem{ozyurt2020effectiveness}
Ozyurt, I.B.: On the effectiveness of small, discriminatively pre-trained
  language representation models for biomedical text mining. In: Proceedings of
  the First Workshop on Scholarly Document Processing. pp. 104--112 (2020)

\bibitem{Ozyurt2021clef}
Ozyurt, I.B.: End-to-end biomedical question answering via bio-answerfinder and
  discriminative language representation models. In: CLEF (Working Notes)
  (2021)

\bibitem{ozyurt2020bio}
Ozyurt, I.B., Bandrowski, A., Grethe, J.S.: Bio-answerfinder: a system to find
  answers to questions from biomedical texts. Database  \textbf{2020} (2020)

\bibitem{pappas2020aueb}
Pappas, D., Stavropoulos, P., Androutsopoulos, I.: Aueb-nlp at bioasq 8:
  Biomedical document and snippet retrieval

\bibitem{peng2016}
Peng, S., You, R., Wang, H., Zhai, C., Mamitsuka, H., Zhu, S.: Deepmesh: deep
  semantic representation for improving large-scale mesh indexing.
  Bioinformatics  \textbf{32}(12),  i70--i79 (2016)

\bibitem{Rae2021clef}
Rae, A., Mork, J., Demner-Fushman, D.: A neural text ranking approach for
  automatic mesh indexing. In: CLEF (Working Notes) (2021)

\bibitem{rae2020automatic}
Rae, A.R., Pritchard, D.O., Mork, J.G., Demner-Fushman, D.: Automatic mesh
  indexing: Revisiting the subheading attachment problem. In: AMIA Annual
  Symposium Proceedings. vol.~2020, p.~1031. American Medical Informatics
  Association (2020)

\bibitem{raffel2019exploring}
Raffel, C., Shazeer, N., Roberts, A., Lee, K., Narang, S., Matena, M., Zhou,
  Y., Li, W., Liu, P.J.: Exploring the limits of transfer learning with a
  unified text-to-text transformer. arXiv preprint arXiv:1910.10683  (2019)

\bibitem{Ribadas2015}
Ribadas, F.J., {De Campos}, L.M., Darriba, V.M., Romero, A.E.: {CoLe and UTAI
  at BioASQ 2015: Experiments with similarity based descriptor assignment}.
  CEUR Workshop Proceedings  \textbf{1391} (2015)

\bibitem{rodriguez2020overview}
Rodriguez-Penagos, C., Nentidis, A., Gonzalez-Agirre, A., Asensio, A.,
  Armengol-Estap{\'e}, J., Krithara, A., Villegas, M., Paliouras, G.,
  Krallinger, M.: Overview of mesinesp8, a spanish medical semantic indexing
  task within bioasq 2020  (2020)

\bibitem{ruas2021}
{Ruas, Pedro and Andrade, Vitor D. T. and Couto, Francisco M.}: {LASIGE-BioTM
  at MESINESP2: entity linking with semantic similarity and extreme multi-label
  classification on Spanish biomedical documents}  (2021)

\bibitem{Sarrouti2021clef}
Sarrouti, M., Gupta, D., Abacha, A.B., Demner-Fushman, D.: Nlm at bioasq 2021:
  Deep learning-based methods for biomedical question answering about covid-19.
  In: CLEF (Working Notes) (2021)

\bibitem{torres2021growth}
Torres-Salinas, D., Robinson-Garcia, N., van Schalkwyk, F., Nane, G.F.,
  Castillo-Valdivieso, P.: The growth of covid-19 scientific literature: A
  forecast analysis of different daily time series in specific settings. arXiv
  preprint arXiv:2101.12455  (2021)

\bibitem{Tsatsaronis2015}
Tsatsaronis, G., Balikas, G., Malakasiotis, P., Partalas, I., Zschunke, M.,
  Alvers, M.R., Weissenborn, D., Krithara, A., Petridis, S., Polychronopoulos,
  D., Almirantis, Y., Pavlopoulos, J., Baskiotis, N., Gallinari, P., Artieres,
  T., Ngonga, A., Heino, N., Gaussier, E., Barrio-Alvers, L., Schroeder, M.,
  Androutsopoulos, I., Paliouras, G.: An overview of the bioasq large-scale
  biomedical semantic indexing and question answering competition. BMC
  Bioinformatics  \textbf{16}, ~138 (2015). \doi{10.1186/s12859-015-0564-6}

\bibitem{tsoumakasBioasq}
Tsoumakas, G., Laliotis, M., Markontanatos, N., Vlahavas, I.: {Large-Scale
  Semantic Indexing of Biomedical Publications}. In: {1st BioASQ Workshop: A
  challenge on large-scale biomedical semantic indexing and question answering}
  (2013)

\bibitem{wang2020cord}
Wang, L.L., Lo, K., Chandrasekhar, Y., Reas, R., Yang, J., Eide, D., Funk, K.,
  Kinney, R., Liu, Z., Merrill, W., et~al.: Cord-19: The covid-19 open research
  dataset. ArXiv  (2020)

\bibitem{Wei2016}
Wei, C.H., Leaman, R., Lu, Z.: {Beyond accuracy: creating interoperable and
  scalable text-mining web services.} Bioinformatics (Oxford, England)
  \textbf{32}(12),  1907--10 (2016). \doi{10.1093/bioinformatics/btv760}

\bibitem{xue2020mt5}
Xue, L., Constant, N., Roberts, A., Kale, M., Al-Rfou, R., Siddhant, A., Barua,
  A., Raffel, C.: mt5: A massively multilingual pre-trained text-to-text
  transformer. arXiv preprint arXiv:2010.11934  (2020)

\bibitem{yang2016learning}
Yang, Z., Zhou, Y., Eric, N.: Learning to answer biomedical questions: Oaqa at
  bioasq 4b. ACL 2016 p.~23 (2016)

\bibitem{yoon2021sequence}
Yoon, W., Jackson, R., Kang, J., Lagerberg, A.: Sequence tagging for biomedical
  extractive question answering. arXiv preprint arXiv:2104.07535  (2021)

\bibitem{you2021bertmesh}
You, R., Liu, Y., Mamitsuka, H., Zhu, S.: {BERTMeSH: deep contextual
  representation learning for large-scale high-performance MeSH indexing with
  full text}. Bioinformatics  \textbf{37}(5),  684--692 (2021)

\bibitem{zavorin2016}
Zavorin, I., Mork, J.G., Demner-Fushman, D.: Using learning-to-rank to enhance
  nlm medical text indexer results. ACL 2016 p.~8 (2016)

\bibitem{Zhang2021clef}
Zhang, Y., Han, J.C., Tsai, R.T.H.: Ncu-iisr/as-gis: Results of various
  pre-trained biomedical language models and logistic regression model in
  bioasq task 9b phase b. In: CLEF (Working Notes) (2021)

\end{thebibliography}

\end{document}